\pdfoutput=1

\documentclass[11pt]{article}

\usepackage[]{acl}

\usepackage{times}
\usepackage{latexsym}

\usepackage[T1]{fontenc}

\usepackage[utf8]{inputenc}

\usepackage{microtype}

\usepackage{tikz}
\usepackage{collcell}
\usepackage{etoolbox}
\usepackage{algorithm}
\usepackage{algorithmic}
\usepackage{booktabs}
\usepackage{nicematrix}
\usepackage{enumitem}
\usepackage{tcolorbox}
\usepackage{enumitem}
\setlist{leftmargin=1mm}
\usepackage{pifont}
\usepackage{setspace}

\usepackage{cleveref}
\crefformat{section}{\S#2#1#3}
\crefformat{subsection}{\S#2#1#3}
\crefformat{subsubsection}{\S#2#1#3}

\title{Generative Data Augmentation using LLMs improves Distributional Robustness in Question Answering}

\author{Arijit Ghosh Chowdhury \\
  University of Illinois Urbana Champaign \\
  \texttt{arijit10@gmail.com} \\\And
  Aman Chadha \\
  Stanford University \\
  Amazon GenAI\thanks{\,\,\,Work does not relate to position at Amazon.} \\
  \texttt{hi@aman.ai} \\}

\begin{document}
\maketitle
\begin{abstract}
Robustness in Natural Language Processing continues to be a pertinent issue, where state of the art models under-perform under naturally shifted distributions. In the context of Question Answering, work on domain adaptation methods continues to be a growing body of research. However, very little attention has been given to the notion of domain generalization under natural distribution shifts, where the target domain is unknown. With drastic improvements in the quality of and access to generative models, we answer the question: How do generated datasets influence the performance of QA models under natural distribution shifts? We perform experiments on 4 different datasets under varying amounts of distribution shift, and analyze how "in-the-wild" generation can help achieve domain generalization. We take a two-step generation approach, generating both contexts and QA pairs to augment existing datasets. Through our experiments, we demonstrate how augmenting reading comprehension datasets with generated data leads to better robustness towards natural distribution shifts.
\end{abstract}

\section{Introduction}


In this work, we perform a systematic study of how "in-the-wild" generation can affect the distributional robustness of question-answering models trained on the popular Stanford Question Answering Dataset (SQUAD) \cite{rajpurkar-etal-2016-squad}. Synthetic data generation is a widely adopted method for domain adaptation in QA systems \cite{shakeri2020end} \cite{yue2021contrastive} \cite{yue2022domain}. However, domain adaptation methods have access to unlabelled/labelled data belonging to the target domain, and do not account for unseen natural distribution shifts. Our work studies the effect of generated data on distribution shifts where the target domain is unseen.


The conception of a dataset has undergone significant evolution in recent times. This transformation has been catalyzed by the advent of generative models trained 'in-the-wild', such as those described in \cite{brown2020language}, \cite{bubeck2023sparks}, and \cite{touvron2023llama}. These models, which use vast and diverse datasets across a range of domains, have facilitated the infusion of the web with synthesized data of high calibre, applicable to an extensive array of conceptual topics. Interestingly, these models are not merely confined to generation based on a pre-established distribution; they possess the capacity for repeated prompting, resulting in the creation of markedly diverse data. In the context of this emerging model paradigm, our research investigates the following query: How do generated datasets affect the distributional robustness of Question Answering models? Specifically, \textbf{natural distribution shifts} in NLP can arise due to differences in the text genre and style, text topics and vocabulary, demographics of the authors, medium of the text (written vs spoken), and other attributes \cite{wang-etal-2022-measure}. A key challenge is that NLP models trained on one data distribution often fail to generalize well to these naturally occurring shifts. For instance, \cite{miller2020effect} found that question answering models experienced average F1 score drops of 3.8 points on news articles, 14 points on Reddit posts, and 17.4 points on Amazon reviews compared to Wikipedia articles. This reveals brittleness of NLP models to natural distribution shifts.

\begin{figure*}[h!]

    \centering
    \includegraphics[width=0.8\textwidth]{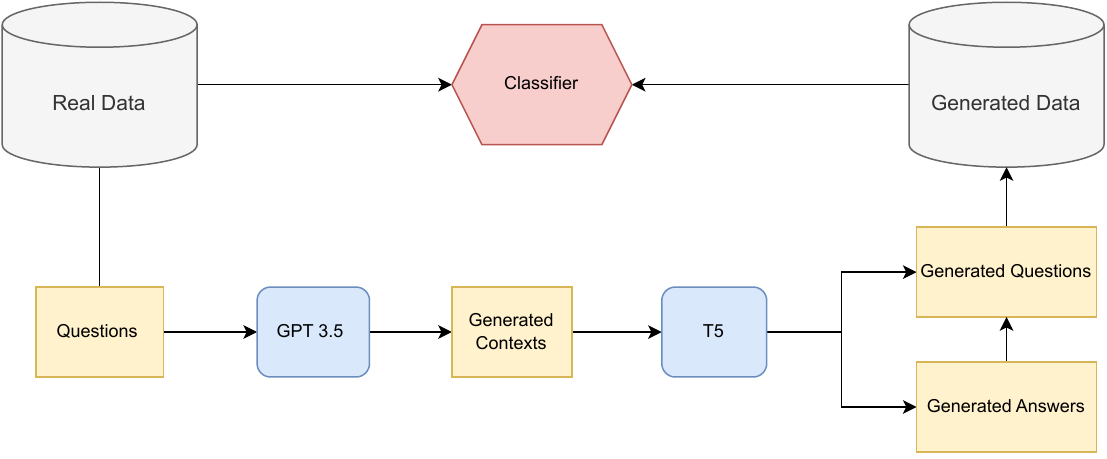}
    \vspace{-2mm}
    \caption{Overview of the generation system. Our method creates a generated dataset which is then augmented with the real dataset to train a question answering model.}
    \label{fig:my_label}
    \vskip -1.1ex
    \label{fig:llm}
\end{figure*}

We present an overview of our generation setup in Figure \ref{fig:my_label}. For generating data, use GPT-3.5 \cite{brown2020language}, and create a question-answering dataset using questions provided in the SQUAD \cite{rajpurkar-etal-2016-squad} dataset. We use a dual generation approach, by first prompting the language model to generate a context for a question given in the SQUAD dataset, and then generating question-answer pairs for the newly generated context. 


Recent surveys, such as \cite{ramponi-plank-2020-neural}, discuss domain adaptation in NLP and divide approaches into \textit{data centric} and \textit{model centric}. We take a data-centric approach, as highlighted by findings from \cite{wang-etal-2022-measure} that demonstrate overlap in test-train data for QA models. The scarcity of research on generalization in QA models, especially with natural distribution shifts, is a motivation for our work, backed by observations from \cite{arora-etal-2021-types} on out-of-distribution data in NLP.

Initial experiments like \cite{longpre-etal-2019-exploration} ventured into domain-agnostic question answering using data augmentation. New datasets introduced by \cite{miller2020effect}, sourced from various platforms, emphasize the effect of natural distribution shifts on QA models. While these studies provide extensive evaluations, our work builds on them by focusing on the impact of large language model (LLM)-generated datasets for QA tasks and further leveraging these datasets for our data augmentation method.

The benefits of generated data have been explored by \cite{gowal2021improving}, showing its potential in adversarial robustness. \cite{bartolo-etal-2021-improving} and \cite{mekala-etal-2022-leveraging} use synthetic and context-generated data respectively for QA and text classification. Our method uses a GPT-3.5 model, as described by \cite{wei2022finetuned}, to generate context for questions. With similar motivations, \cite{bansal2023leaving} demonstrates the application of Stable Diffusion in diverse dataset creation for image tasks.

\vspace{-1mm}
\begin{tcolorbox}
[colback=blue!5!white,colframe=blue!75!black,title=\textsc{Our Contributions}]
\vspace{-2mm}
\begin{itemize}
[leftmargin=1mm]
\setlength\itemsep{0em}
\begin{spacing}{0.85}
    \item[\ding{224}] {\footnotesize 
    {\fontfamily{phv}\fontsize{8}{9}\selectfont
    We propose a framework to improve the distributional robustness of reading comprehension models in the presence of natural distribution shifts.}
    }

    \item[\ding{224}] {\footnotesize 
    {\fontfamily{phv}\fontsize{8}{9}\selectfont
    Through a thorough quantitative evaluation, we evaluate the capabilites of LLMs to generate high quality synthetic data for question answering tasks. }
    }    

\vspace{-5mm}    
\end{spacing}    
\end{itemize}
\end{tcolorbox}

\section{Methodology}

\subsection{Context Generation}

We first generate contexts by conditioning it on a question present in the SQUAD dataset. This allows the language model to generate a paragraph that can be used to generate question-answer pairs. Since the paragraph is generated using an existing question, the generated context is consistent with the informative trivia format of SQUAD-like datasets. We also ensure that the generated contexts are diverse yet complimentary to the original dataset, as highlighted by \cite{gowal2021improving}. To maintain further consistency, the generated context is clipped to be within 250 words, based on the average context length present in the SQUAD dataset. We prompt GPT 3.5 (gpt-3.5-turbo) \footnote{https://platform.openai.com/docs/models} in the following manner: \textit{Generate a paragraph which answers the following question: (question) }. Here the question is sampled from the SQUAD dataset. Figure \ref{fig:llm} demonstrates the generation process. Additionally, the Appendix \ref{sec:appendix} contains examples from the generation process.

\subsection{Question Answer Generation}

After the context is created, the generated paragraph is used to create question-answer pairs. This is done by using a T5 based question generation model \cite{lopez2020TransformerbasedEQ} that is trained on the SQUAD dataset, which takes a paragraph has an input and returns a question-answer pair. We use the open source\footnote{https://github.com/patil-suraj/question-generation} implementation for this model. Additionally we also filter out QA pairs based on round-trip consistency \cite{alberti2019synthetic}.

\section{Experiments}

\begin{table*}[]
\begin{centering}
\resizebox{0.8\textwidth}{!}{%
\begin{tabular}{@{}lcccccccccl@{}}
\toprule
\textbf{Dataset}      & \multicolumn{2}{c}{SQUAD} & \multicolumn{2}{c}{NewWiki} & \multicolumn{2}{c}{NYT} & \multicolumn{2}{c}{Amazon} & \multicolumn{2}{c}{Reddit} \\ \midrule
\textbf{Metrics}      & F1          & EM          & F1           & EM           & F1         & EM         & F1           & EM          & F1           & EM          \\ \midrule
Real data             & 90.4        & 83.0        & 89.4         & 79.2         & 86.4       & 76.1       & 79.9         & 66.4        & 80.1         & 67.1        \\
Generated data        & 79.5        & 64.6        & 80.1         & 65.3         & 76.5       & 63.2       & 72.4         & 59.5        & 72.7         & 60.2        \\
Real + Wiki-samples        & 93.4        & 85.2        & 89.3         & 77.3         & 79.4       & 78.1       & 76.4         & 66.6        & 78.8         & 63.2        \\

Real + Generated data & \textbf{92.7}        & \textbf{84.7}        & \textbf{91.1}         & \textbf{80.4}         & \textbf{88.9}       & \textbf{79.3}       & \textbf{80.3}         & \textbf{67.1}        & \textbf{81.7}         & \textbf{68.7}        \\ \bottomrule
\end{tabular}
}
\caption{Generated datasets demonstrate robustness to natural distribution shifts.}\label{tab:res1}
\end{centering}
\end{table*}

\begin{table*}[]
\begin{centering}
\resizebox{0.8\textwidth}{!}{%
\begin{tabular}{@{}lcccccccccl@{}}
\toprule\textbf{Dataset}      & \multicolumn{2}{c}{SQUAD} & \multicolumn{2}{c}{NewWiki} & \multicolumn{2}{c}{NYT} & \multicolumn{2}{c}{Amazon} & \multicolumn{2}{c}{Reddit} \\ \midrule
\textbf{Metrics}      & F1          & EM          & F1           & EM           & F1         & EM         & F1           & EM          & F1           & EM          \\ \midrule
Real + 50\% Generated data& 91.4        & 81.1        & 90.4         & 82.2         & 87.4       & 77.1       & 79.7         & 65.4        & 80.3         & 67.4        \\
Real + 100\% Generated data        & 92.7        & 84.7        & 91.1         & 80.4         & \textbf{88.9}       & \textbf{79.3}       & 80.3         & 67.1        & \textbf{81.7}         & \textbf{68.7}        \\
Real + 200\% Generated data & \textbf{92.9}        & \textbf{84.8}       & \textbf{91.3}         & \textbf{80.7}         & 88.5       & 79.1       & \textbf{80.9}         & \textbf{67.3}        & 80.8         & 68.1        \\ \bottomrule
\end{tabular}
}
\caption{Performance on varying amounts of data. Using equal measures of real and generated data is essential. }\label{tab:res2}
\end{centering}
\end{table*}

\begin{table*}[]
\begin{centering}
\resizebox{0.8\textwidth}{!}{%
\begin{tabular}{@{}lcccccccccl@{}}
\toprule
\textbf{Dataset}      & \multicolumn{2}{c}{SQUAD} & \multicolumn{2}{c}{NYT} & \multicolumn{2}{c}{Amazon}  \\ \midrule
\textbf{Metrics}      & F1          & EM          & F1           & EM           & F1         & EM         \\ \midrule
Real data             & 90.4        & 83.0        & 86.4         & 76.1         & 79.9           & 66.4              \\
Real + Generated data (Questions Only)        & 91.5        & 82.7        & 85.7               & 75.6       & 77.4         & 63.5             \\
Real + Generated data (Contexts + Questions)  & \textbf{92.7}        & \textbf{84.7}        & \textbf{88.9}         & \textbf{79.3}         & \textbf{80.9}       & \textbf{67.3}                    \\ \bottomrule
\end{tabular}
}
\caption{Ablation Study demonstrating how context generation is key to robustness.}\label{tab:abl}
\end{centering}
\end{table*}

\subsection{Setup}
We train an extractive reading comprehension model using SQUAD V1.1, using the RoBERTA-Base model across all our experiments. We use a learning rate of $3e-5$, a batch size of $16$ and run our experiments for $3$ epochs each. We use the implementation provided by HuggingFace, and run our models on a stand-alone Nvidia A100 GPU provided by Google Colab. We do not use GPT-3.5 as a baseline since the purpose of this study is to specifically measure the performance by smaller models.

For all our experiments, we measure F1 and Exact Match scores to quantify performance on Natural Distribution Shift (NDS) datasets.

\subsection{Datasets}

We use the following datasets created by \cite{miller2020effect} to set up our testbed:

    The \textbf{New Wikipedia} dataset contains newer QA pairs from wikipedia articles used by the SQUAD V1.1 dataset. Contains 7,938 test samples from 48 contexts.
    The \textbf{New York Times} dataset contains articles from New York times which are then used to annotate QA pairs in the same format as SQUAD. It is ensured that the passage length statistics stay the same. Contains 10,065 test samples from 46 articles.
    \textbf{Reddit} dataset contains articles from Reddit where the authors concatenated each post’s title with its body. This dataset contains 9,803 test samples from 1,969 posts.
    The \textbf{Amazon Product Reviews} dataset contains user generated product reviews from the "Home and Kitchen" category on Amazon. This data contains 9,885 test samples from 1,909 reviews.

    \section{Results}

\subsection{Does generated data help with distributional robustness?}

We evaluate the F1 and Exact Match scores of models trained with different datasets on natural distribution shifts (NDS) benchmarks. We note the average EM and F1 numbers across three random seeds in Table \ref{tab:res1}. The models are trained on an equal amount of real and generated data.


We find that the model, when trained on SQUAD, when subjected to natural distribution shift datasets, the model's performance significantly deteriorates. A noteworthy observation was that exclusive training on the generated data resulted in substandard performance on both the SQUAD and its Natural Distribution Shift (NDS) datasets. The inferior absolute performance could be potentially attributed to the distribution disparity between the source and the generated training datasets. Interestingly, we observe that for the model trained on the generated data, the performance gaps on the real validation dataset and its NDS datasets are low, which might be attributed to the benefits of training on diverse generated data. This highlights the contributions of the generated data in improving robustness, as opposed to simply generating more data for training. 

We also sample paragraphs from Wikipedia and generate questions from those paragraphs, instead of letting GPT3.5 generate the paragraphs. This improves in-domain performance on SQUAD, but leads to drops in performance across out of domain datasets, further emphasizing on the effectiveness of the in-the-wild context generation on distribution shifts.

Finally, we expose our model to an evenly-distributed blend of real and generated datasets, with the goal of investigating the impact of generative augmentations. Our results reveal that the absolute performance of the model, when trained with a combination of real and generated data, either parallels or exceeds the performance of models trained exclusively on either real or generated datasets, across all naturally distributed datasets. This observation suggests that the incorporation of real data into the training process is indeed essential for attaining superior absolute performance.

To summarize, while using solely generated data improves robustness at the expense of absolute performance, a blend of real and artificially generated data presents the ideal balance for robust and precise training.

\subsection{How much generated data is needed?}

Here, we investigate how different combinations of the generated dataset can help
the classifiers take advantage of the complementary strengths of the two data sources (Table \ref{tab:res2}). 

To do so, we assessed the average performance of models trained
with three different input mixing combinations created by using 50\%, 100\%, and 200\% of the generated dataset. 
We observed an increase in performance on shifted datasets as the size of the generated
data increases while keeping the amount of real data fixed. However, when the proportion of the generated data increases twofold while keeping the proportion of the real data fixed, we observe that the performance gains are only marginal. Additionally, we note that using only half of the generated data does not provide enough meaningful signal in terms of diversity and does not lead to major performance improvements compared to training  on real data.

Overall, we found that the ideal split between real and generated data is a 50-50 split where the two datasets are able to compliment each other, in terms of providing both diversity and in-domain samples at the same time.

\subsection{Is context generation needed?}

Table \ref{tab:abl} demonstrates the importance of generating both contexts and questions for improving model robustness to distribution shifts. When only questions are generated for existing contexts, performance on the original SQuAD dataset improves slightly, while performance degrades substantially on the out-of-distribution NYT and Amazon datasets. This indicates that generating questions alone overfits models to the SQuAD distribution, reducing robustness. In contrast, generating both contexts and questions leads to consistent improvements in performance across all datasets. The dual generation approach enhances model robustness by exposing the model to more diversity during training, leading to better generalization. The results demonstrate that generating varied contexts in addition to targeted question generation is crucial for improving robustness to natural distribution shifts, rather than question generation alone.

\section{Conclusion and Future Avenues}

We created a framework that enhances the performance of reading comprehension models by supplementing real datasets with a diverse dataset generated by contemporary, real-world generative models. Our findings indicate that this training method yields superior results on test datasets and those with natural distribution shifts, due to the added robustness from training on the generated data as opposed to traditional methods.  
In the future, we want to explore a more extensive comparison against question generation methods and how this paradigm fits into fine-tuning larger models.

\bibliography{anthology,custom}

\appendix

\newpage
\section*{Frequently Asked Questions (FAQs)}\label{sec:FAQs}

\begin{itemize}
[leftmargin=4.5mm]
\setlength\itemsep{1.5em}
    \item[\ding{93}] {\fontfamily{lmss} \selectfont \textbf{How are we sampling questions to generate paragraphs?}}
    \vspace{-2mm}
    \begin{description}
    \item[\ding{224}] One question is sampled per context in the original SQUAD dataset to condition the paragraph generation.
    \end{description}

    \item[\ding{93}] {\fontfamily{lmss} \selectfont \textbf{Why don't we generate new contexts from scratch?}}
    \vspace{-2mm}
    \begin{description}
    \item[\ding{224}] This is done to create topically consistent datasets, and run a controlled experiment where only determine whether LLM generated contexts provide linguistic and stylistic diversity.
    \end{description}

    \item[\ding{93}] {\fontfamily{lmss} \selectfont \textbf{Why don't we prompt GPT-3/3.5/4 to generate QA pairs as well?}}
    \vspace{-2mm}
    \begin{description}
    \item[\ding{224}] In our attempts to do so, we realised that GPT prompts do not generate outputs that stay faithful to the extractive nature of the SQUAD dataset. Furthermore, we wanted to limit our API usage.
    \end{description}

\end{itemize}    

\onecolumn

\appendix

\newpage

\renewcommand{\thesubsection}{\Alph{section}.\arabic{subsection}}
\renewcommand{\thesection}{\Alph{section}}
\setcounter{section}{0}

\section{Appendix: Example Generations}
\label{sec:appendix}

This section provides supplementary material in the form of additional examples
to bolster the reader's understanding of the concepts presented in this work.

\begin{table}[H]
\begin{centering}
\resizebox{\textwidth}{!}{%
\begin{tabular}{p{1.5in}|p{6in}}

\toprule
\textcolor{red}{Original Context}   & The American bison is a large bovid which inhabited much of western North America prior to the 1800s, living on the prairies in large herds. However, the vast herds of bison attracted market hunters, who killed dozens of bison for their hides only, leaving the rest to rot. Thousands of these hunters quickly eliminated the bison herds, bringing the population from several million in the early 1800s to a few hundred by the 1880s. Conservation efforts have allowed the population to increase, but the bison remains near-threatened.                                                                                                                                                                                                                                                                                                                                                                                                                                                                                                                                                                                                                                                                                                                                                                                                                                          \\ \midrule
\textcolor{red}{Sampled Question}   & "What type of creature is the American bison?"                                                                                                                                                                                                                                                                                                                                                                                                                                                                                                                                                                                                                                                                                                                                                                                                                                                                                                                                                                                                                                                                                                                                                                                                                                                                                                                                                   \\ \midrule
\textcolor{red}{Prompt}            & Generate a paragraph that answers the following question: "What type of creature is the American bison?"                                                                                                                                                                                                                                                                                                                                                                                                                                                                                                                                                                                                                                                                                                                                                                                                                                                                                                                                                                                                                                                                                                                                                                                                                                                                                                                                                  \\ \midrule
\textcolor{red}{Generated Context}  & The American Bison, often colloquially referred to as buffalo, is a North American species of bison that once roamed across the vast expanses of the continent in large herds. These colossal creatures are mammals, belonging to the family Bovidae, which includes other large, hoofed mammals such as domestic cattle and goats. Adult bison are imposing creatures; they are known to stand up to 6.5 feet tall at the shoulder and can weigh over a ton, making them the heaviest land animals in North America.They have a robust body structure with a large, muscular hump over the shoulders and a shaggy, dark brown winter coat which gets lighter in the summer. Their large heads sport a pair of short, sharp horns, used both for defense and in fights for status within the herd. Despite their bulk, bison are surprisingly agile and fast, capable of reaching speeds up to 35 mph.The American Bison plays a crucial role in maintaining the health and biodiversity of the prairie ecosystem.  \\ \midrule
\textcolor{red}{Generated Questions} & \textbf{Q}: To which species does the term "buffalo" colloquially refer in North America? \textbf{A}: American Bison, \textbf{Q}: How does the American Bison use its horns? \textbf{A}: for defense and in fights for status within the herd                                                                                                                                                                                                                                                                                                                                                                                                                                                                                                                                                                                                                                                                                                                                                                                                                                                                                                                                                                                                                                                                                                                                                                                        \\ \bottomrule
\end{tabular}

}
\end{centering}
\caption{Example \#1 -- Context, Question, and Answer Generation.}
\end{table}

\begin{table}[H]
\begin{centering}
\resizebox{\textwidth}{!}{%
\begin{tabular}{p{1.5in}|p{6in}}

\toprule
\textcolor{red}{Original Context}   & Exhibitions and annual horse shows in all districts and a national horse and cattle show at Lahore are held with the official patronage. The national horse and cattle show at Lahore is the biggest festival where sports, exhibitions, and livestock competitions are held. It not only encourages and patronises agricultural products and livestock through the exhibitions of agricultural products and cattle but is also a colourful documentary on the rich cultural heritage of the province with its strong rural roots.                                                                                                                                                                                                                                                                                                                                                                                                                                                                                                                                                                                                                                                                                                                                                                                                                                        \\ \midrule
\textcolor{red}{Sampled Question}   & "What is Punjab's biggest festival?"                                                                                                                                                                                                                                                                                                                                                                                                                                                                                                                                                                                                                                                                                                                                                                                                                                                                                                                                                                                                                                                                                                                                                                                                                                                                                                                                                  \\ \midrule
\textcolor{red}{Prompt}            & Generate a paragraph that answers the following question: "What is Punjab's biggest festival?"                                                                                                                                                                                                                                                                                                                                                                                                                                                                                                                                                                                                                                                                                                                                                                                                                                                                                                                                                                                                                                                                                                                                                                                                                                                                                                                                                    \\ \midrule
\textcolor{red}{Generated Context}  & Punjab's biggest festival is Baisakhi, also known as Vaisakhi, celebrated with great fervor and enthusiasm in the region. It's a significant festival for the Punjabi people and is observed annually on the 13th or 14th of April. Baisakhi is not only a harvest festival, marking the time when farmers reap their rabi (winter) crops, but it also has deep religious significance, particularly for Sikhs. It commemorates the formation of the Khalsa Panth, a Sikh political and military group, by Guru Gobind Singh Ji in 1699. The day is marked by visiting gurdwaras, participating in processions, dancing the traditional Bhangra and Gidda, and sharing meals in the community. The joyous festival is a celebration of culture, heritage, and communal unity, encapsulating the vibrant spirit of Punjab. \\ \midrule
\textcolor{red}{Generated Questions} & \textbf{Q}: When is Baisakhi observed? \textbf{A}: annually on the 13th or 14th of April, \textbf{Q}: Why is Baisakhi significant for farmers? \textbf{A}:  a harvest festival, marking the time when farmers reap their rabi (winter) crops                                                                                                                                                                                                                                                                                                                                                                                                                                                                                                                                                                                                                                                                                                                                                                                                                                                                                                                                                                                                                                                                                                                                                                                      \\ \bottomrule
\end{tabular}

}
\end{centering}
\caption{Example \#2 -- Context, Question, and Answer Generation.}
\end{table}

\begin{table}[H]
\begin{centering}
\resizebox{\textwidth}{!}{%
\begin{tabular}{p{1.5in}|p{6in}}

\toprule
\textcolor{red}{Original Context}   & In Canada, the Royal Newfoundland Constabulary was founded in 1729, making it the first police force in present-day Canada. It was followed in 1834 by the Toronto Police, and in 1838 by police forces in Montreal and Quebec City. A national force, the Dominion Police, was founded in 1868. Initially the Dominion Police provided security for parliament, but its responsibilities quickly grew. The famous Royal Northwest Mounted Police was founded in 1873. The merger of these two police forces in 1920 formed the world-famous Royal Canadian Mounted Police.                                                                                                                                                                                                                                                                                                                                                                                                                                                                                                                                                                                                                                                                                                                                                                                                                                        \\ \midrule
\textcolor{red}{Sampled Question}   & "When did Canada get its first police?"                                                                                                                                                                                                                                                                                                                                                                                                                                                                                                                                                                                                                                                                                                                                                                                                                                                                                                                                                                                                                                                                                                                                                                                                                                                                                                                                                 \\ \midrule
\textcolor{red}{Prompt}            & Generate a paragraph that answers the following question: "When did Canada get its first police?"                                                                                                                                                                                                                                                                                                                                                                                                                                                                                                                                                                                                                                                                                                                                                                                                                                                                                                                                                                                                                                                                                                                                                                                                                                                                                                                                                    \\ \midrule
\textcolor{red}{Generated Context}  & The first policing institution in Canada was established in the early 19th century. The Royal Newfoundland Constabulary, founded in St. John's, Newfoundland in 1729, is considered the oldest civil law enforcement body in North America. However, it primarily served as a paramilitary unit, dealing with the city's disorderly sailors and soldiers. Later, the North-West Mounted Police (NWMP), which is often seen as the direct predecessor to the current Royal Canadian Mounted Police (RCMP), was formed on May 23, 1873. The establishment of the NWMP marked a significant shift in the approach to law enforcement in Canada, transitioning from a military-style model to a community-focused approach. The NWMP, and later the RCMP, played a crucial role in establishing the rule of law across Canada, particularly in the west and north during the periods of rapid expansion and development in the late 19th and early 20th centuries. \\ \midrule
\textcolor{red}{Generated Questions} & \textbf{Q}: What is the oldest civil law enforcement body in North America? \textbf{A}: The Royal Newfoundland Constabulary, \textbf{Q}: Which organization is seen as the direct predecessor to the Royal Canadian Mounted Police? \textbf{A}:  the North-West Mounted Police                                                                                                                                                                                                                                                                                                                                                                                                                                                                                                                                                                                                                                                                                                                                                                                                                                                                                                                                                                                                                                                                                                                                                                                     \\ \bottomrule
\end{tabular}

}
\end{centering}
\caption{Example \#3 -- Context, Question, and Answer Generation.}
\end{table}

\begin{table}[H]
\begin{centering}
\resizebox{\textwidth}{!}{%
\begin{tabular}{p{1.5in}|p{6in}}

\toprule
\textcolor{red}{Original Context}   & The Earth of the early Archean (4,000 to 2,500 million years ago) may have had a different tectonic style. During this time, the Earth's crust cooled enough that rocks and continental plates began to form. Some scientists think because the Earth was hotter, that plate tectonic activity was more vigorous than it is today, resulting in a much greater rate of recycling of crustal material. This may have prevented cratonisation and continent formation until the mantle cooled and convection slowed down. Others argue that the subcontinental lithospheric mantle is too buoyant to subduct and that the lack of Archean rocks is a function of erosion and subsequent tectonic events.                                                                                                                                                                                                                                                                                                                                                                                                                                                                                                                                                                                                                                                                                                                                                                                                                                        \\ \midrule
\textcolor{red}{Sampled Question}   & "During what time period was the Archean era?"                                                                                                                                                                                                                                                                                                                                                                                                                                                                                                                                                                                                                                                                                                                                                                                                                                                                                                                                                                                                                                                                                                                                                                                                                                                                                                                                                \\ \midrule
\textcolor{red}{Prompt}            & Generate a paragraph that answers the following question: "During what time period was the Archean era?"                                                                                                                                                                                                                                                                                                                                                                                                                                                                                                                                                                                                                                                                                                                                                                                                                                                                                                                                                                                                                                                                                                                                                                                                                                                                                                                                                    \\ \midrule
\textcolor{red}{Generated Context}  & The Archean Eon, one of the four principal eons of Earth's geologic time scale, occurred approximately 4.0 to 2.5 billion years ago. It follows the Hadean Eon and precedes the Proterozoic Eon. The Archean era is divided into four geologic time periods, known as eras: the Eoarchean, Paleoarchean, Mesoarchean, and Neoarchean. This time period was crucial in Earth's history as it saw the formation of the planet's crust and the development of the earliest forms of life. Despite the extreme conditions, with high temperatures and a lack of oxygen, early life forms such as cyanobacteria began to emerge, which eventually played a role in shaping the Earth's atmosphere by producing oxygen through photosynthesis. \\ \midrule
\textcolor{red}{Generated Questions} & \textbf{Q}: What eon follows the Hadean Eon and precedes the Proterozoic Eon?\textbf{A}: The Archean Eon, \textbf{Q}: Despite what conditions did early life forms begin to emerge during the Archean Eon? \textbf{A}: the extreme conditions, with high temperatures and a lack of oxygen                                                                                                                                                                                                                                                                                                                                                                                                                                                                                                                                                                                                                                                                                                                                                                                                                                                                                                                                                                                                                                                                                                                                                                                     \\ \bottomrule
\end{tabular}

}
\end{centering}
\caption{Example \#4 -- Context, Question, and Answer Generation.}
\end{table}

\end{document}